\documentclass[letterpaper]{article}
\usepackage{aaai}
\usepackage{times}
\usepackage{helvet}
\usepackage{courier}
\usepackage{times}  
\usepackage{helvet}  
\usepackage{courier}  
\usepackage{url}  
\usepackage{graphicx}  
\usepackage{booktabs} 
\usepackage{amsmath}
\usepackage{amssymb}

\usepackage{tabularx}
\usepackage{multirow}
\usepackage{graphicx}
\usepackage{float}
\usepackage{url}
\usepackage{xcolor}
\usepackage{url}
\usepackage{color, colortbl}
\usepackage{enumitem}
\usepackage[utf8]{inputenc}
\usepackage{balance}
\usepackage{xfrac}
\usepackage{amssymb}
\usepackage[utf8]{inputenc}
\usepackage{graphicx}

\usepackage{algorithm}
\usepackage{amsmath}
\usepackage[noend]{algpseudocode}

\usepackage{caption}
\usepackage{subcaption}

\usepackage[T1]{fontenc}
\usepackage{graphicx}
\begin{document}

%
\title{Optimized Tracking of Topic Evolution}

\author{Patrick Kiss\textsuperscript{1}, Elaheh Momeni\textsuperscript{2}\\
\textsuperscript{1}eMentalist, Vienna, Austria\\
\textsuperscript{2}University of Vienna, Vienna, Austria}
\maketitle
\begin{abstract}
\begin{quote}
Topic evolution modeling has been researched for a long time and has gained considerable interest. A state-of-the-art method has been recently using word modeling algorithms in combination with community detection mechanisms to achieve better results in a more effective way. We analyse results of this approach and discuss the two major challenges that this approach still faces. Although the topics that have resulted from the recent algorithm are good in general, they are very noisy due to many topics that are very unimportant because of their size, words, or ambiguity. Additionally, the number of words defining each topic is too large, making it difficult to analyse them in their unsorted state. In this paper, we propose approaches to tackle these challenges by adding topic filtering and network analysis metrics to define the importance of a topic. We test different combinations of these metrics to see which combination yields the best results. Furthermore, we add word filtering and ranking to each topic to identify the words with the highest novelty automatically. We evaluate our enhancement methods in two ways: human qualitative evaluation and automatic quantitative evaluation. Moreover, we created two case studies to test the quality of the clusters and words. In the quantitative evaluation, we use the pairwise mutual information score to test the coherency of topics. The quantitative evaluation also includes an analysis of execution times for each part of the program. The results of the experimental evaluations show that the two evaluation methods agree on the positive feasibility of the algorithm. We then show possible extensions in the form of usability and future improvements to the algorithm.
\end{quote}
\end{abstract}

\section{Introduction}
Modeling large temporal text corpora is key to understanding how topics attract attention and social and cultural norms evolve over time. Such evolutions are especially prevalent in social systems, where the rapid exchange of ideas can quickly change the  importance of  topics  and the attention they receive.\\
While embedding methods are promising as a diachronic tool, they are limited to modeling only words or documents and they are not well-suited to model changes due to the stochastic nature of their training. This implies that models trained on exactly the same data could produce vector spaces where words have the same nearest neighbours but do not have the same coordinates. An alternative approach to this topic modeling task is to use the different extensions of Dynamic Topic models  (DTM)~\cite{Blei:2006}, which use a Bayesian technique to infer the
topic structure in the corpora of documents. DTM perceives a document as a mixture of a small number of topics, and topics as a (relatively sparse) distribution over word types. While documents may indeed be seen as a mixture of topics, documents can still be expected to be semantically coherent. However, this prior preference for semantic coherence is not encoded in the model, and any such observation
of semantic coherence found in the inferred topic distributions is in some sense stochastic. 
\\
There has been a new method that has proved to work well \cite{Elaheh}, overcoming the limitations of available solutions. \cite{Elaheh} uses the word embedding algorithm to create a network of words based on input texts and then applies community detection algorithms to find topics. The main advantages of this method are: (i) instead of using embedding to model changes of words it works with clusters of words as topics, which are more meaningful and have higher interpretability, (ii) for modeling the evolution it avoids the issue of the alignment of embedding models and change detection by using embedding models directly, (iii) it can accelerate the modeling process by parallelizing the process in different snapshots, and (iv) it can model change and evolution in different forms (grow, contract, merge, split, birth, die, and survive). 
\\
While the newest approach achieved excellent results, several challenges have remained, resulting in the user needing to handpick clusters to define topic evolution at a later time. These challenges are:

\begin{itemize}
    \item \emph{Challenge 1: large number of topics:}
    On the one hand, when looking at the topics resulting from the system \cite{Elaheh}, it was noticeable that there was a vast number of topics - up to 100 for the given dataset, many of which seemed uninteresting. We found two reasons for that: first, while some topics had only a few words, other topics seemed to have many words, meaning that they did not seem to form one coherent topic, and second, topics did not appear to be important in general since they defined a topic that was very general in itself. The goal here must be to find a method to describe the importance of a topic using a metric and filtering the data based on that metric.

    \item \emph{Challenge 2: Large number of words in topics:}
    Another noticeable thing was the high amount of words that were found to be in most topic clusters. It is highly unlikely that all of these words, 270,000, actually contribute to a topic. This was especially noticeable in a dataset with a considerable number of texts, like the legal dataset used for the showcase in \cite{Elaheh}. They made the manual selection process for their showcase more difficult as there was more unnecessary data to look through.
\end{itemize}

In this paper, following a short overview of the recently published algorithm, we propose two approaches to overcome the challenges mentioned above. Firstly, we combine two different cluster analysis metrics, namely the centrality and the cluster frequency, to filter unwanted topics. Secondly, we use k-core decomposition in combination with the TFIDF to filter and rank words in a cluster. In addition to these proposed approaches, we also discuss other variants or combinations that we have tried but that have not proven to be useful for our task. Furthermore, performing empirical experiments we investigated when and with which setup the proposed approaches can perform effectively and efficiently considering user requirements and characteristics of the dataset. These results enable further development of the approach as a customized and adaptive solution.
\\
Finally, using two historical corpora, legal data and robotic news, spanning two languages (English and German), we demonstrate that our approach with a short execution time is able to detect dynamic and interesting topics with high coherency over the years. To do this, we utilized both qualitative and quantitative evaluation methods. For the quantitative evaluation method, we use a large external word dataset to evaluate coherency of proposed topics by our system using PMI values for each topic. Additionally, we evaluate the run time and possible reasons for spikes in run time. For the qualitative analysis of the result, we asked multiple users to look at the data and answer questions regarding the topic coherency and word ranking of the clusters. We also created a case evaluation detailing results for two different topics in both datasets.

\section{Background}
As noted in the introduction, researchers have been very interested in the area of dynamic topic detection, especially regarding topics in very dynamic environments, for example in social media. Researchers are attempting to detect conversation triggers \cite{Pres2} and find out what is trending \cite{pres3} and what the dynamics of these trending topics are \cite{Pres1}.

\subsubsection{Dynamic Topic Modeling}
\cite{Blei:2006} took state of the art approaches for static topic modeling using Latent Dirichlet Allocation (LDA) and proposed an algorithm to support a whole time series of documents.\\
The biggest difference between the state of the art static topic models and their proposed dynamic topic models is that with static topic models, documents always include one of the same pools of topics, whereas with dynamic topic modeling, topics evolve and may not always include the same words, same types of words or same number of words.\\
Their base assumption is that a document consists of a certain number of topics and each topic consists of a certain number of words.\\
Furthermore, they define a set of topics in a timestamp $t$ as the evolution of the topics from timestamp $t_{-1}$. Their algorithm essentially works by using two multinomial distributions. One distribution defines a topic per document and another defines the words for a topic (as is done in static topic modeling). Then the algorithm uses mean parameterization to evolve these distributions over time.\\
The main disadvantage of this approach is that it considers words, but does not take into consideration dynamic topic events, as illustrated in the next chapter

\subsubsection{Topic Evolution Modeling using Word Embeddings}
\cite{Elaheh} uses the word2vec word embedding modeling technique to create a model to analyze the evolution of topics.\\
They use embedding models to get contextual information from text corpora (one embedding model per timestamp) and use the information from these models to get topics (clusters) using a dynamic clustering method and similarity networks. Additionally, they can detect specific events for a topic, using the same event types as \cite{Ilhan:2015}.\\
In their experiments, they used the skip-gram model to generate the word embedding models of the two given test sets, and then generated the semantic similarity networks for each snapshot using the well-known Louvain community detection algorithm to get the needed clusters of words. Finally, they analyzed the events happening to those clusters over time.\\
In their evaluations, they compared their novel algorithm to the already existing alternative approach of Dynamic Topic Models (DTM) \cite{Blei:2006}. The main problem of the existing approaches (like DTM) is that they do not try to detect the different events that could happen to topics over time, such as the ones explained in \cite{Ilhan:2015}. In their evaluation, \cite{Elaheh} examined the coherency of topics found by their method and the capability to detect dynamic topics in a short execution time. They illustrated the evolution of a topic, including a set of detected events as described in \cite{Ilhan:2015}.\\
To test the topic coherency, they compared and scored the results of the DTM approach and their new approach, resulting in better results of the new approach throughout the tests. Furthermore, they asked several human test subjects to rate the results (topics) to see if they considered the detected topics as useful. This test showed that the human testers widely agreed with the scoring from the above test.\\ To show the new method was able to detect dynamic topics, they showed a graph as an example that showed the dynamic evolution of the topic "race" in Figure 1 of their paper. \cite{Elaheh}\\
Finally, they pointed out the dramatic difference between the above  method and the one used previously as regards execution time. Their approach is much faster overall, and it could be made even faster by parallelizing the computations for each time step. For such large data sets as theirs, the DTM approach could not even compute the results in time (for the previous tests they had to reduce the number of snapshots as the execution time of the DTM was too long).\\
While this state-of-the-art algorithm \cite{Elaheh} had good results, automatically processing them was not immediately possible as i) the topic collection was really noisy, there being many topics with only a few words resulting in up to 100 topics for each timestamp and ii) the topics that contained many words had so many words that a way of ranking these was necessary. In fact, the biggest number of clusters we found in one of their datasets was 270,000. Examples in their dataset include clusters that consist of only four words or clusters that consist of tens of thousands of words

\section{Proposed Approach}
Following, we first give a short overview of the recent algorithm discussed above \cite{Elaheh}, then two approaches to overcome the aforementioned challenges are proposed.
\subsection{Topic Evolution Modelling Method}
The system receives a set of temporal text documents, $D^{1}, ...,D^{n}$ and for each snapshot $D^{i}$, returns a set of topics $T^{i}_1, ...,T^{i}_m$.
The \emph{p}th topic of the \emph{i}th  snapshot, $T^{i}_p$, is assumed to be a semantically coherent set of
words and consists of a tuple of three: $T^{i}_p = <e^{i}_p; T^{i+1}; W^{i}_p>$.
Each topic contains an event (labeled from a set of events: Survive, Grow, Merge, Split, Contract, and Die) with regard to the later snapshot, a set of relevant topics from a later snapshot, $T^{i+1}$, and a set of semantically related words, W. The first two attributes are useful for tracking the evolution of topics. The third attribute shows the content of the topic. 
\\
\textbf{Learning Semantic Space:} The distributional methods learn a semantic space that maps words to a continuous vector space $\mathbb{R}^{s}$, where s is the dimension of vector space. A family of neural language methods embeds words in a fixed-dimension vector space in such a way that words in similar contexts tend to produce similar representations in a vector space. These methods project words in a lower-dimensional vector space, so that each word $w_i$ is represented by an s-dimensional vector $v_i$. We use the word2vec method~\cite{Mikolov:2013,Mikolov2:2013} to learn word vector representation (word embeddings) that we track across time.
\\
\textbf{Modeling Dynamic Topics:} For modeling topics we utilize a clustering-based approach on a semantic representation network, extracted from trained embedding models. 
\\
Formally, a network N(W, E) denotes the sets of words
W and edges E in a network. Edges show semantic similarity between words. 
Evolving network N is described over a time period and it will be decomposed into a sequence of static snapshots $N^{1}, ...,N^{n}$. 
\\
For each snapshot, $t$ we list a set of all words $W^{t}$ and develop $N^{t}$ by assessing semantic similarity between a pair of words leveraging trained embedded vectors. The edge between two words is created when the similarity score between their embedded vectors are higher than the \emph{semantic similarity threshold, $\varphi$}. The semantic similarity is calculated based on the cosine similarity score, formally:

\begin{equation}
\scriptstyle{
simSemantic(v^{t}_i,v^{t}_j)=\frac{v^{t}_i,v^{t}_j}{{||v^{t}_i||}_2{||v^{t}_j||}_2} > \varphi
}
\end{equation}
\\
Next, the topic detection leverages community detection algorithms (i.e. network clustering approach). While $N^{t}$ represents the \emph{t}th snapshot of the network, $T^{t}=\{T^{t}_1,T^{t}_2,...,T^{t}_m\}$ represents the set of topics in the snapshot t.\\
Finally, according to the available approach~\cite{Aris:2016,Ilhan:2015} for dynamic community detection, we use a matching metric in order to track evolution of a topic from one snapshot to the following snapshots and measure the similarity between two clusters in successive time steps. Two topics match if their similarity value
$sim(T^{t}_i,T^{t+1}_j)$ exceeds a user set \emph{similarity threshold $\theta$}, formally:

\begin{equation}
\scriptstyle{
sim(T^{t}_i,T^{t+1}_j)=min \bigg( \frac{|{T^{t}_i\bigcap T^{t+1}_j}|}{|T^{t}_i|}, \frac{|{T^{t}_i\bigcap T^{t+1}_j}|}{|T^{t+1}_j|}\bigg) > \theta
}
\end{equation}
\\
Being able to reuse traditional community detection techniques without having to modify them for detecting clusters in each snapshot separately is one of the main advantages of this described solution. Furthermore, this method results in parallelizing and accelerating community detection dramatically on all snapshots, thus reducing running time. However, this solution is not perfect due to the instability of community detection algorithms. The majority of community detection algorithms that work well are stochastic, meaning two runs on the same network do not necessarily provide the same partition. The solution can sometimes provide different results for two networks which are the same apart from tiny modifications. An alternative approach with higher stability would be to study all snapshots  simultaneously~\cite{Tantipathananandh:2007}.
\\
\textbf{Modeling Topic Evolution:} 
In order to model evolution of topics, we use an existing metric, namely \emph{Instability}, in order to compute the percentage of increase/decrease in the number of words in a topic~\cite{Ilhan:2015}. More formally, given a topic $T^{t}_i$ has $n^{t}_i$ members at time snapshot $t$
and a successor cluster $T^{t+1}_j$ has $n^{t+1}_j$ members at time snapshot $t+1$, the \emph{Instability} is defined as: $inst(T^{t}_i,T^{t+1}_j)=(\sfrac{n^{t+1}_j}{n^{t}_i}) -1$. 
\\
Next, considering a user-defined \emph{instability threshold, ($\phi$)}, six events proposed by Ilhan et al.~\shortcite{Ilhan:2015}, including Grow, Survive, Contract, Split, Merge and Die are identified to model the evolution of the topics. More precisely, after positive matching between $T^{t}_i$ and $T^{t+1}_j$, we could then label the similar topic as being one of the following topic evolution types:

\begin{itemize}[noitemsep,nolistsep]
\item \textbf{Contract}: If $inst(T^{t}_i,T^{t+1}_j) < {-} {\phi}$. The topic has been contracted (i.e. there is a substantial percentage decrease in the number of members). 
\item \textbf{Survive}: If $ {-}{\phi} < inst(T^{t}_i,T^{t+1}_j) < \phi$. The topic has survived (i.e. there is a negligible increase/decrease in the number of members). 
\item \textbf{Grow}: If $inst(T^{t}_i,T^{t+1}_j) > \phi$. The topic has grown (i.e. there is a substantial percentage increase in the number of members). 
\item \textbf{Split}: A topic $T^{t}_i$ at time t may match with a set of topics $T^{t+1}_*= \{{T^{t}_i}...{T^{t+1}_j}\}$ in a later snapshot in a split case. 
\item \textbf{Merge}: A set of topics $T^{t}_*= \{{T^{t}_1}...{T^{t}_i}\}$ may match to a topic $T^{t+1}_j$ in the subsequent snapshot $t + 1$ in a  merge case. 
\item \textbf{Die}: If there is no similar topic at a later snapshot, this means $\theta$ is not exceeded. Then it is assumed that the topic dies.
\end{itemize}

\subsection{Optimization 1: Reduction of proposed topics}

When looking at the topics resulting from the proposed approach above, it is noticeable that for a large corpus there will be a vast number of topics - up to 100 for a given dataset - many of which seemed uninteresting. We found two reasons for that: first, while some topics had only a few words, other topics seemed to have many words, meaning that they did not seem to form one coherent topic, and second, topics did not appear to be important in general since they defined a topic that was very general in itself. The goal here must be to find a method to describe the importance of a topic using network metrics and filtering the data based on that metric.
Leveraging common network metrics like density, centrality, or more, we can tackle this issue and optimize our algorithm.
While the centrality tells us about how important a node is inside a network by means of external connections to the node, the density tells us how dense a network is, meaning to what extent the nodes connect.
\\
With regard to our investigation of different combinations of network analysis mechanisms (see \emph{Density-Centrality combination} tests sub-section), we decided to develop a system that uses the centrality and cluster frequency to describe the "importance" of a cluster and the sparseness of its nodes, and adapts a simple term frequency for clusters. That allows us to categorize the clusters to make a customized selection based on a users' parameters.\\

\textbf{Algorithm 1} The pseudocode below (Algorithm 1) describes the two main steps necessary to define a topic as "important" or "unimportant"
\begin{enumerate}
    \item The first function finds the time series for each topic. In an earlier step, we calculated the similarities between topics of two timestamps. Using these for a greedy algorithm is an easy way to find connections over different time stamps. As the next topic in the time series, we define the topic in the next timestamp that has the highest similarity value. Similarly, the previous topic is defined by the topic in the last time stamp that has the biggest similarity value.
    \item The second function describes how we use the information we have just gathered to filter our clusters. We use our two values to first check if each given topic satisfies the thresholds set by a user. If it does not, we check the whole time series until we either find that the topic satisfies all thresholds in at least one timestamp, or that it is an unimportant topic.
\end{enumerate}
\begin{algorithm}
\caption{Cluster Filtering Algorithm}\label{euclid}
\begin{algorithmic}[1]
\Function{getTimeSeries}{$cluster\_dict$, $similarity\_values\_path$}
\State GET all time stamps available in the dataset
\State SORT timestamps from lowest to highest
\For{every timestamp}
    \For{every cluster in the timestamp}
        \If{timestamp is not the last timestamp in the series}
            \State GET next timestamp from list of timestamps
            \State LOAD cluster similarity matrix for the two timestamps
            \If{the two clusters have similarity values}
                \State SET the most significant match to be the next cluster
                \State SET the current cluster as the previous cluster in the biggest match
            \EndIf
        \EndIf
    \EndFor
\EndFor
\EndFunction
\Function{isClusterUsable}{$cluster\_dict$, $cluster\_number$, $\gamma$, $\theta$}
\If{size of cluster > $\alpha$}
    \If{centrality > $\gamma$ \textbf{AND} cluster frequency > $\theta$}
        \State \textbf{return} True
    \Else
        \State GET the time series of clusters of the current cluster
        \For{each cluster in the time series}
            \If{centrality > $\gamma$ \textbf{AND} cluster frequency > $\theta$}
                \State \textbf{return} True
            \EndIf
        \EndFor
    \EndIf
\EndIf
\EndFunction
\end{algorithmic}
\end{algorithm}
\emph{\textbf{Importance by constraints.}} The algorithm represents the following rules:
\begin{itemize}
    \item \emph{\textbf{Rule 1:}} The number of words $n$ in a cluster must be higher than the set cluster size threshold $\alpha$: $n > \alpha $
    \item \emph{\textbf{Rule 2:}} The centrality $c$ of a cluster must be higher than the set centrality threshold $\gamma$: $c > \gamma$
    \item \emph{\textbf{Rule 3:}} The cluster frequency $cf$ and/or density $d$ of a cluster must be higher than the set cluster frequency threshold $\theta$ or density threshold $\delta$: $cf > \theta$ and/or $d > \delta $
    \item \emph{\textbf{Rule 4:}} If Rules 2 and/or 3 fail, check if they are true for the other clusters in the clusters time series: $T \epsilon \{T_{t1} ... T_{tn}\}$
\end{itemize}

\begin{figure}[h]
\includegraphics[width=4cm]{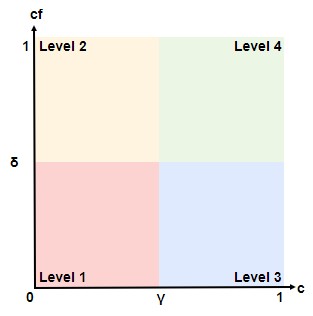}
\centering
\caption{The overview of the four types of importance, where the threshold on the x-axis is the centrality threshold $\gamma$ and the threshold on the y-axis is the cluster frequency threshold $\theta$}
\label{fig:quadrant_overview}
\end{figure}
\begin{figure*}
\centering
    \begin{subfigure}[b]{0.20\textwidth}
            \includegraphics[width=\linewidth]{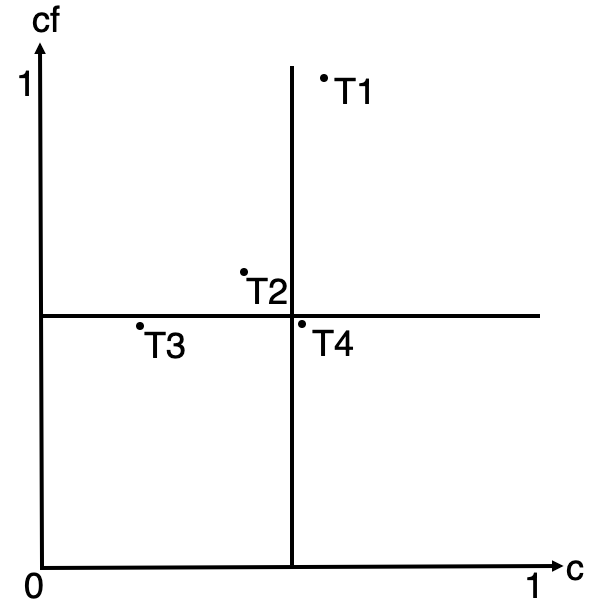}
            \caption{t0}
    \end{subfigure}%
    \begin{subfigure}[b]{0.20\textwidth}
            \includegraphics[width=\linewidth]{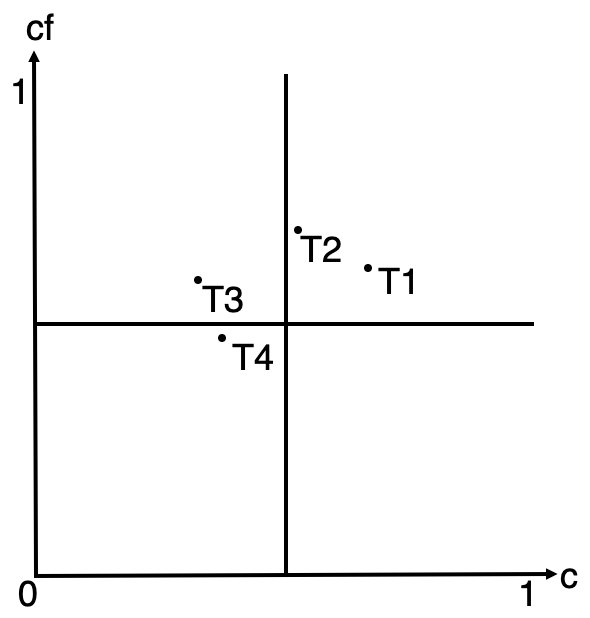}
            \caption{t1}
    \end{subfigure}%
    \begin{subfigure}[b]{0.20\textwidth}
            \includegraphics[width=\linewidth]{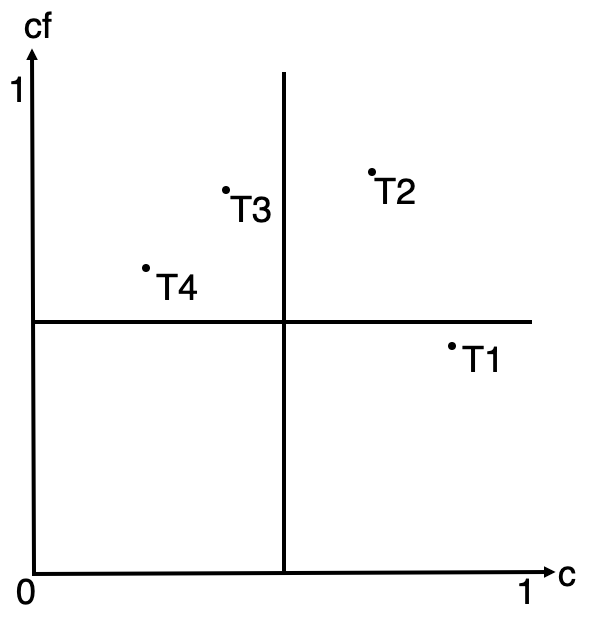}
            \caption{t2}
    \end{subfigure}%
    \begin{subfigure}[b]{0.20\textwidth}
            \includegraphics[width=\linewidth]{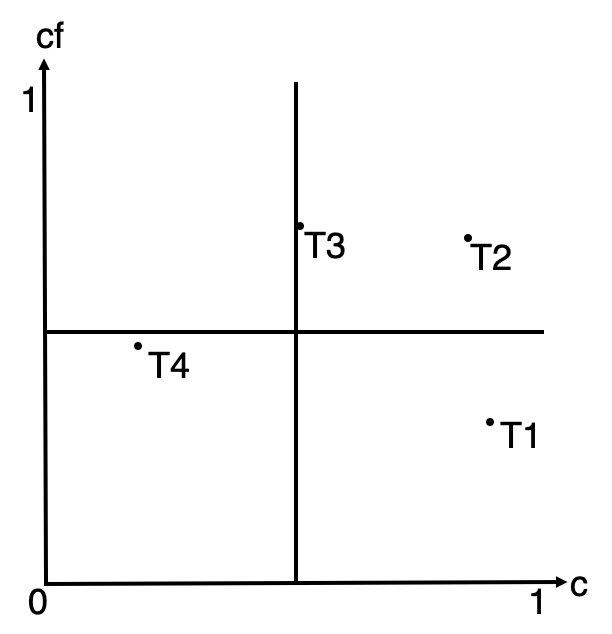}
            \caption{t3}
    \end{subfigure}
    \caption{The evolution of 4 topics from timestamps t0 to t3 shown with centrality (c) and cluster frequency (cf)}
    \label{fig:customized_clusters_centrality_frequency}
\end{figure*}

Considering centrality and cluster frequency, four levels of importance are defined for each topic. By selecting different values for these two metrics, centrality and cluster frequency, the end-user  can adaptively reduce the number of topics for final output of the system.  Figure \ref{fig:quadrant_overview} and \ref{fig:customized_clusters_centrality_frequency} (a fictitious mock up) illustrate the functionality of this algorithm with some fictitious topics changing their importance over time, defined by their centrality and frequency values. The four quadrants represent the four different levels of importance:
\begin{enumerate}
    \item \emph{\textbf{Level 1 (Red)}}:\\
    The quadrant in the lower left contains all the topics that seem to be very unimportant. Their words are neither used very much nor are the words in them very central to the text.
    \item \emph{\textbf{Level 2 (Yellow)}}:\\
    The quadrant in the upper left contains all the topics that have words with many occurrences, but they are still not used in connection with other topics very much. This means they still have a low centrality, which is low importance.
    \item \emph{\textbf{Level 3 (Blue)}}:\\
    This quadrant contains topics that have very central words that are often used in connection with other topics but are more unique in the dataset.
    \item \emph{\textbf{Level 4 (Green)}}:\\
    This quadrant has the most important topics that have both a high cluster frequency, signaling much use of words that define a topic, and a high centrality.  These topics are also important over the whole network of a timestamp.
\end{enumerate}
Our algorithm checks these values in all timestamps for each topic. We define a topic as important if it at any time moves into Level 4. This considers the evolution of a topic where a topic seems to be of low interest over some timestamps, but becomes more important eventually, thus making the topic interesting. Such a topic then stays in the system. Figure \ref{fig:customized_clusters_centrality_frequency} shows the state of 4 topics, T1 to T4, at timestamps t0 to t3:

\begin{itemize}
    \item \textbf{T1} has been losing importance over time, resulting in a decrease in use of its words. It is observable that the importance falls over time. The algorithm would still keep this topic as important as it was at some time (in this case two timestamps) situated in the 4th Level of our threshold legend. 
    \item \textbf{T2}, on the other hand, rises in importance and moves up mostly in terms of the centrality value. It is considered as an important topic by the algorithm as it mostly resides in the upper right threshold area.
    \item \textbf{T3} is a slow-moving topic that slowly becomes more important over time. Even though it barely enters the area of the chart we defined as the most important, it is kept by the algorithm.
    \item \textbf{T4} is a topic that always resides in the rather unimportant parts of our definition of "importance," and accordingly would be ignored by our algorithm.
\end{itemize}

\subsubsection{Density - Centrality combination tests}
Before finding the algorithm based on centrality and cluster frequency, we tested the combination of centrality and density.\\
While centrality and density certainly are very interesting values to analyze for a cluster in terms of defining cluster importance, density did not add any significant value to our importance calculation. \cite{STOKMAN200110509} also note that density is not a good measure to compare networks. While it can analyze connections between words inside a topic, it is not useful to consider and compare the density of a cluster to the one of another cluster. Taking this into consideration, we needed another value for our clusters that made sense to use when comparing topics. We chose cluster frequency, which we defined as the mean of all word frequencies inside a topic, and then normalized the resulting values, so all cluster frequency values are between 0 and 1.\\
Having this third value in the system allows us to interchange the density value with the cluster frequency value. Nevertheless, it also enables us to convert it into a three-dimensional threshold problem where we use all three values: centrality, density, and the cluster frequency.\\
Analogous to the above example, we also have illustrations providing an example overview of this algorithm that uses the centrality in combination with the cluster frequency. The four charts in Figure \ref{fig:customized_clusters_centrality_frequency} show the state of 4 topics T1 to T4 at timestamps t0 to t3.
\begin{itemize}
    \item \textbf{T1} is situated in the lower left part of the graph and has never had any big movement towards better values. This means it will be marked as skipped.
    \item \textbf{T2} moves into the upper right part of the graph at timestamp t2, and is marked as important since it was above the thresholds at some point.
    \item \textbf{T3} stays in the upper right quadrant for most of the time we have data from, so it is marked as important too.
    \item \textbf{T4} has been moving lower in density and stays with low centrality value over time. It will be marked as unimportant.
\end{itemize}
\begin{figure*}
\centering
\begin{subfigure}[b]{0.20\textwidth}
                \includegraphics[width=\linewidth]{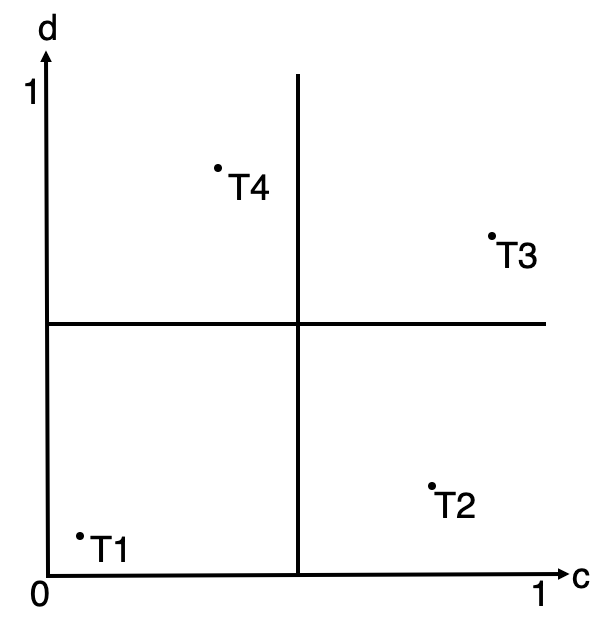}
                \caption{t0}
        \end{subfigure}%
        \begin{subfigure}[b]{0.20\textwidth}
                \includegraphics[width=\linewidth]{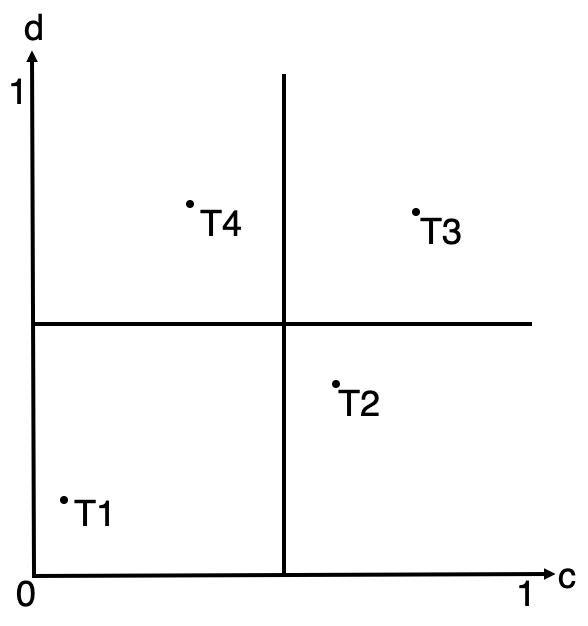}
                \caption{t1}
        \end{subfigure}%
        \begin{subfigure}[b]{0.20\textwidth}
                \includegraphics[width=\linewidth]{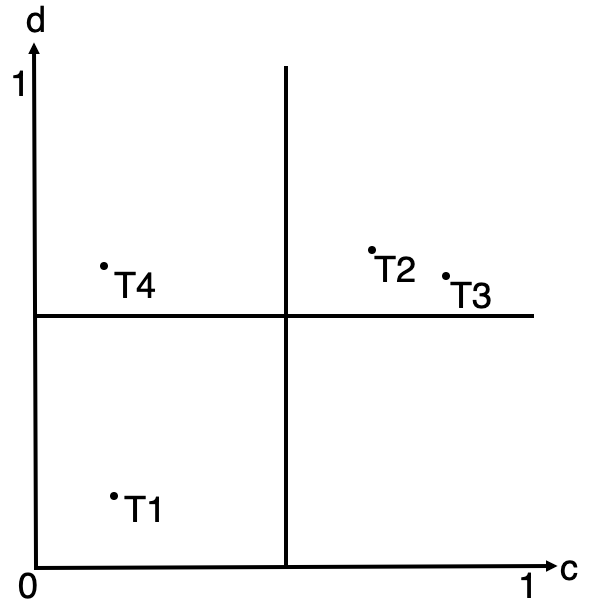}
                \caption{t2}
        \end{subfigure}%
        \begin{subfigure}[b]{0.20\textwidth}
                \includegraphics[width=\linewidth]{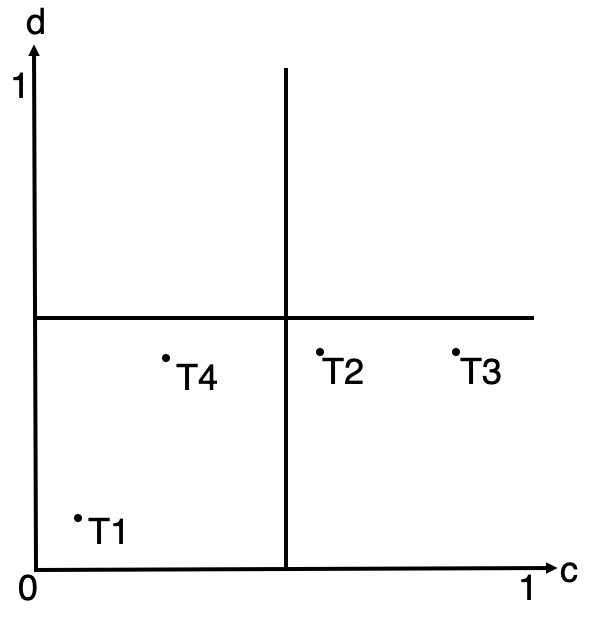}
                \caption{t3}
        \end{subfigure}
        \caption{The evolution of 4 topics from timestamps t0 to t3 shown with centrality (c) and density (d)}
        \label{fig:customized_clusters_centrality_density}
\end{figure*}

\subsection{Optimization 2: Reduction of words considering importance and novelty}

Our solution to this challenge again leverages the combination of two methods: the k-core-decomposition and the TF-IDF. When running this algorithm, the question was asked if it would not be enough only to use TF-IDF ranking and skip all k-core decomposition calculations. While the TF-IDF mechanism itself generates a good ranking, we concluded that adding the k-core decomposition is worthwhile. By first applying k-core decomposition, we receive all words in the topic that are used in a very dense way. When re-adding the words the TF-IDF ranks very high, we also guarantee that domain-specific words with high novelty stay in the topic.\\

\textbf{Algorithm 2}
The pseudocode of Algorithm \ref{euclid} below describes the main steps necessary to reduce words in out topics.\\
\begin{algorithm}
\caption{Word Reduction Algorithm}\label{euclid}
\begin{algorithmic}[1]

\Function{reduceWords}{$graph$, $core\_number$, $tfidf\_data$}
\State CALCULATE mean TF-IDF value
\State REDUCE words in graph by core number
\State GET list of words deleted from graph
\For{every deleted word}
    \If{word is in TF-IDF data}
        \State GET TF-IDF value of word
        \If{TF-IDF value > mean TF-IDF value}
            \State ADD word back to the graph
        \EndIf
    \EndIf
\EndFor
\Return{reduced graph}
\EndFunction
\end{algorithmic}
\end{algorithm}
\\
\emph{\textbf{Importance by rules.}} When looking at the above algorithm, the following rules for the topic reduction can be defined:
\begin{itemize}
    \item \emph{\textbf{Rule 1:}} Reduce the word network to only keep words in or above the core number defined by the user
    \item \emph{\textbf{Rule 2:}} Re-add all nodes that were reduced from the network, but have a higher TF-IDF than the mean TF-IDF
\end{itemize}
\textbf{Using the k-core decomposition to reduce the number of words in a topic.}
 In order to reduce the number of words in our clusters, we use the network structure we already have. We chose to use the k-core decomposition algorithm. Using this method, we receive the innermost important words in the network that are often used and clustered together. While the k-core decomposition is an efficient way to reduce network sizes and less-involved nodes, considering it as a direct clustering mechanism instead of the Louvain algorithm was not in question. Due to the nature of how k-core decomposition works, finding topic clusters from a network of words is not possible since it would only result in one cluster per timestamp.\\
Our initial tests showed that the given network structure did not allow for core numbers higher than 10.
When taking a closer look, it quickly became visible that using the k-core mechanism alone would not lead to the best result as it does not qualify as an efficient ranking algorithm. If we only ranked by core values, we would have many words that are in the inner core, but we could not tell any difference between those words.\\
Table \ref{table:k_core_overview} shows an example distribution of the words among k-cores for one cluster. As the table shows, most of them are set in the very high core numbers which led us to add an extra ranking mechanism to the algorithm.\\
\begin{table}[h]
\centering
\begin{tabular}{ |l|l| }
\hline
Core Number & Number of words in core \\
\hline
    0 & 0 \\
    1 & 206 \\
    2 & 1482 \\
    3 & 3014 \\
    4 & 4276 \\
    5 & 4841 \\
    6 & 4813 \\
    7 & 4742 \\
    8 & 2753 \\
    9 & 1000 \\
    10 & 0 \\
\hline
\end{tabular}
\caption{Distribution of words amongst cores}
\label{table:k_core_overview}
\end{table}
\begin{figure}
    \centering
    \includegraphics[width=.40\textwidth,keepaspectratio]{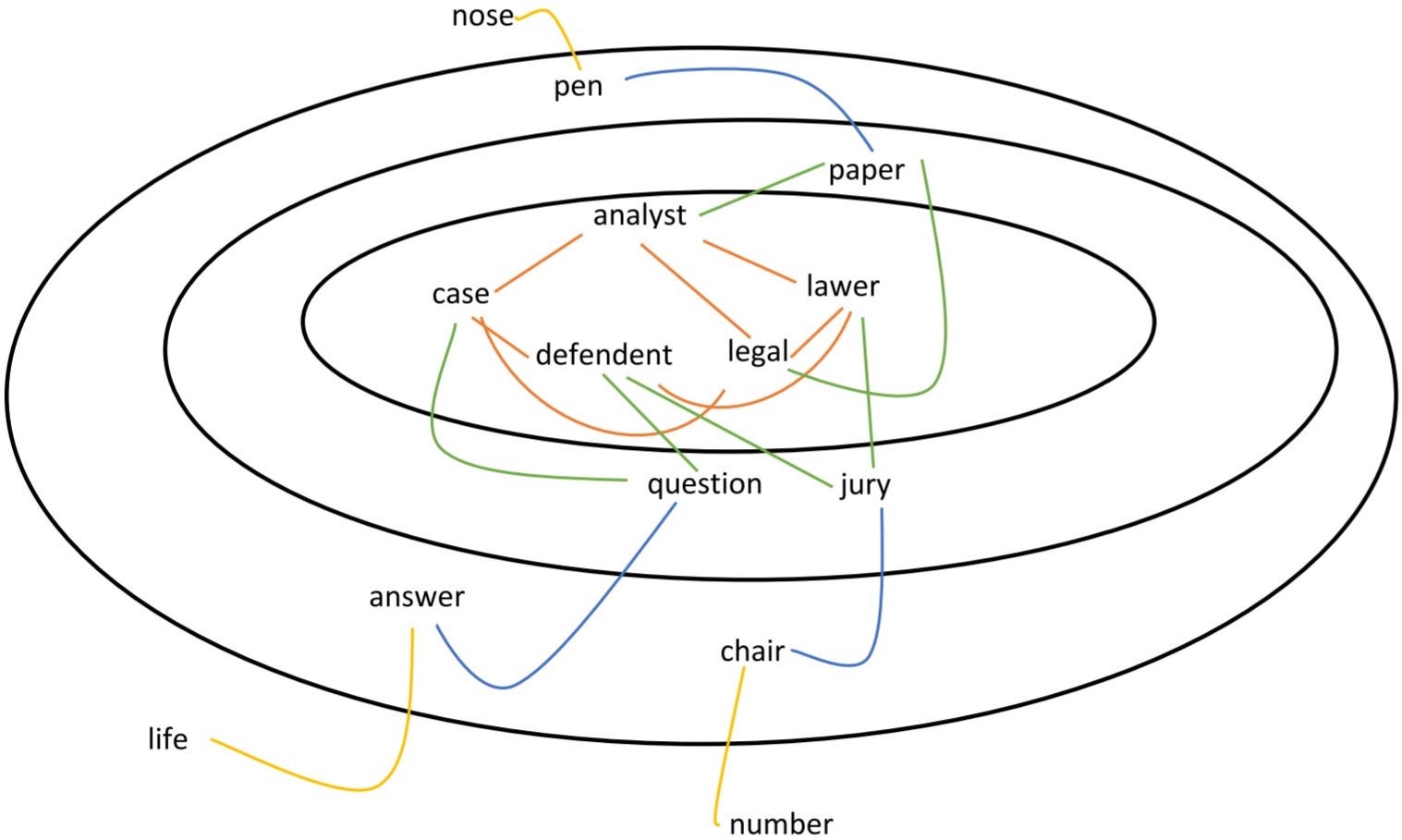}
    \caption{An example mockup of k-core clusters in the legal data domain.}
    \label{fig:k_core_mockup}
\end{figure}
Figure \ref{fig:k_core_mockup} shows an example of what a k-core clustering for the legal domain could look like. The densest and most often used words are in the central cluster. The lesser-used words are at the outside border of a graphic and have a lower core number. The core number here corresponds to the node degree of the single nodes. The inner cluster has a core number of 4, and the outer core, which is not in a cluster anymore as its "cluster" is the whole graph, has a core number of one.\\
\textbf{Adding a TF-IDF ranking for keeping novel terms.}
As a ranking mechanism, we chose to use the well known TF-IDF - algorithm. We added a calculation method that calculates the TF-IDF on a per time stamp basis during the model generation. This was necessary as we later did not process any documents themselves. Instead, we always worked on a per time stamp basis. Therefore, we needed to calculate the TF-IDF values for each word on a per time stamp basis as well, not on a per-document basis like it is usually calculated.\\
At the moment of word reduction, we apply the k-core decomposition algorithm to receive the core values for each of our words and limit the list of words to the words contained in the top n cores, as defined by the user (see the customization section). Next, we implement a "safeguard" that also adds back any words in the network that have a higher TF-IDF than the mean TF-IDF of the whole timestamp to the list. This is to assure that very novel words that would have been reduced by the k-core mechanism but can still be very important are not missed. Lastly, we rank the words in the list to have the most important words.\\
The evaluation section explores some of the clusters and the ranking of words from real a life, unstructured data example.\\

\subsection{Customization}
The thresholds for the proposed system are completely customizable for the end-user. Ideally, it would be better to create a more user-friendly algorithm that did not have as many inputs as this one does, but it is nearly impossible to find suitable values and numbers for thresholds that we could pre-setup for every user. Either the required thresholds are dependent on the domain of the data, or the desired outcome of a user is simply different (for example, the degree of reduction depends on whether the user wants to reduce many words or only a few). While the whole system has many input variables, we now focus on the ones that are important for this algorithm. 
\begin{itemize}
    \item \emph{\textbf{Similarity threshold $\sigma$.}} The similarity threshold tells us how dense we want our networks to be. It is used at network generation where we ask the models for the similarity values between all words. With the threshold, we only add edges between words that have a similarity that is bigger than the set threshold. This bigger threshold results in having a smaller network of words.
    \item \emph{\textbf{Centrality threshold $\gamma$.}} The user can set a centrality threshold between 0 and 0.9. The higher this threshold is set, the more clusters will be filtered by the algorithm. A threshold of 0 would mean that all topics are kept, while a threshold of 0.9 would result in only the top topics staying in the system. While it is possible to set a value of 1 for this threshold, it is improbable that any topics would be kept since that would need a fully connected network.
    \item \emph{\textbf{Density Threshold $\delta$.}} The density threshold, just like the centrality threshold, can be set to values between 0 and 0.9. The user can limit the density value a cluster must have to limit the selection of clusters to the ones that are more interconnected.
    \item \emph{\textbf{Top similar words threshold $\tau$.}} This is another input parameter in order to limit the network size. It defines how many of the top similar words are chosen to be added to the network. However, it must be noted that the similarity threshold is only applied after this threshold, so the number of words used can be significantly lower if there are not any words that are very similar.
    \item \emph{\textbf{Cluster frequency threshold $\theta$.}} The cluster frequency threshold is a value between 0 and 0.9 to limit the cluster frequency (the averaged term frequency of words in a cluster). Only topics whose cluster frequency value is above the threshold set by the users are marked important. In order to achieve comparable values throughout datasets, the cluster frequency is normalized to values between 0 and 1.
    \item \emph{\textbf{Cluster size threshold $\alpha$.}} The cluster size threshold is an integer number and is used to ignore all clusters that have less than $\alpha$ nodes.
\end{itemize}

For the second optimization algorithm, a user can define the lowest core number they want to choose words from. Depending on the core number entered by the user, this results in only the most used words being added to the final list presented to a user, or results in a huge number of words.\\
Apart from that, this part of the algorithm is straight forward, and no more customization is necessary.

\section{Experimental Evaluation}

\subsection{Dataset}  
We evaluate the performance of our proposed approach on two corpora: a Legal Dataset and Web News articles related to Robotics News and social media posts.\\
\textbf{\emph{Dataset1-Legal Dataset:}} A collection of legal opinions from court websites and from data donations between 1920 to
2014 in English\footnote{Available at \url{https://www.courtlistener.com/opinion/}}. It contains $2,902,806$ documents. We divided this collection into 19 snapshots. Each snapshot is related to a period of 5 years (such as 1920-1925 or 1925-1930).\\
\textbf{\emph{Dataset2-Robotics Dataset:}}
The robotics dataset is a collection of texts gathered from the web and includes social media platforms, news articles, and crowdfunding pages about robotics, which contains $204,521$ documents\footnote{Available at \url{https://ementalist.ai}}. For each post we have the date of publication. This means we can sort them by date and use them for the trending topics evolution tracking algorithm. The data was provided as a mongodb collection.

\subsection{Setup}
For each snapshot of each dataset, first, we trained embedding models using the skip-gram implementation of the Gensim library. Second, using trained models, we created semantic similarity networks for each snapshot with a satisfactory density level by setting $\varphi = 0.5$. Third, we used Louvain, a well-known community detection algorithm, to obtain clusters of words. Finally, starting from the earliest snapshot of each dataset, we identified dynamic topics and their related evolution events.\\
In order to demonstrate the effectiveness and accuracy of our proposed optimization approaches, we utilized both qualitative and quantitative evaluation methods. For the qualitative analysis of the result, we asked multiple users to look at the data and answer questions regarding the topic coherency and word ranking of the clusters. We also created a case evaluation detailing results for two different topics in both datasets. For the quantitative evaluation method, we use a large external word dataset to evaluate coherency of proposed topics from our system using a PMI value for each topic.\\
Additionally, we evaluate the run time and possible reasons for spikes in run time. Finally, in order to compare the performance of our approach against available topic modeling approaches, we trained DTM models on both datasets using DTM Gensim implementation (choosing the default hyper-parameters)\footnote{As the execution time of DTM for all snapshots is very long (see Execution Time Evaluation Section), we trained the models using only the latest four snapshots of each dataset.}.

\subsubsection{\emph{Time series creation:}}
To get a better idea about the cluster evolution, we stored a singular object that was needed to analyze cluster evaluation. This object makes it easy to follow a topic's evolution and gather interesting information on a clusters time series. A script looks at all the previously calculated similarity values and stores this information on a per time-series basis. Additionally, each time series is given a unique identifier for identification purposes.

\subsubsection{\emph{Pairwise mutual information score calculation:}}
For the Pairwise mutual information (PMI) score evaluation, as described above, multiple scripts were necessary:
\begin{enumerate}
    \item \emph{Random Cluster Selection:}The first setup for the PMI evaluation was necessary to gather random clusters. Here it was important for us to choose random clusters from different setups, given as a parameter for this script. The setup for the PMI evaluation will be described in the evaluation section.
    \item \emph{Google Ngram Counting: }The second script utilized the capabilities of the google\_ngrams\_downloader module. The script loads multiple ngram files at the same time to calculate the frequencies of word pairs as well as all single words occurring in those word pairs. After a whole file was read, the frequencies are stored in a json file. All the json file results are located in the appendix.
    \item \emph{Google Ngram Counting for Large Files: }The third script is used for huge n-gram files (we used it for files with more than 200K lines). We often ran into a connection problem with files needing a long time to process the script described above. To circumvent this, we downloaded those files and wrote this script for the analysis of local files. Additionally, the script was parallelized to enable faster processing, where each process was counting the word pairs for a specific part of the file, and adding all of them up after the script finished.
    \item \emph{Ngram Count Aggregation: } The fourth script for ngram processing is tasked with adding all results from the 723 5-gram files together to have one main json file telling us all the frequencies from all word pairs and singular words.
    \item \emph{PMI value calculation: }The last script takes the frequencies of the word pairs in the 5-gram dataset and calculates the PMI value for the different clusters the word pairs are from. The result is again stored in a json file.
\end{enumerate}

\subsubsection{\emph{Execution time:}}
To evaluate the execution time, we tracked the time of each step of our program for all tests and stored them. Later, we wrote a script that takes those times and combines them for each step so we could have a sense of which combinations take a long time and which run faster.
\subsubsection{\emph{Use cases development:}}
For the Use case evaluation, one script was necessary that lets us follow the main keywords for a topic through the time steps. It has a list of words as input and goes through all clusters to check for these keywords. Its output is a json formatted file that shows us for each timestamp which keyword was found in which cluster. 

\subsection{Results of the Qualitative Evaluation}
\subsubsection{Topic Coherency:}
\begin{figure}
    \centering
    \includegraphics[width=.50\textwidth,keepaspectratio]{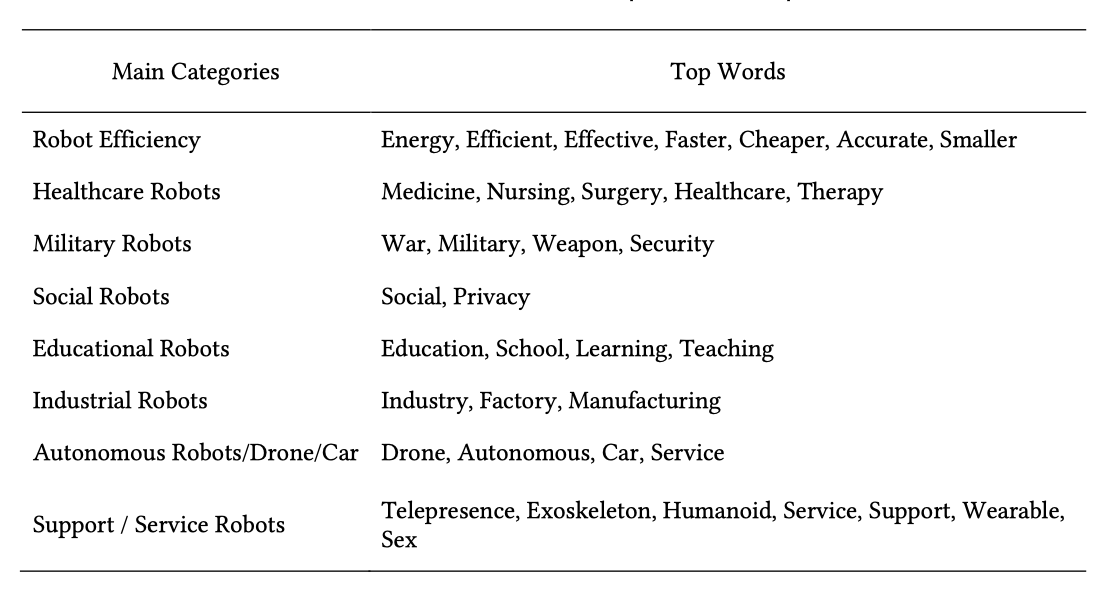}
    \caption{Table of robotics topic and their top words \cite{robot_paper}.}
    \label{fig:robo_topics}
\end{figure}

\begin{figure}
    \centering
    \includegraphics[width=.40\textwidth,keepaspectratio]{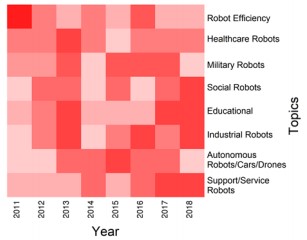}
    \caption{Heatmap for topic importance over time \cite{robot_paper}.}
    \label{fig:robo_heatmap}
\end{figure}

For the subjective analysis of topics, we used our algorithm on the robotics dataset and provided the data to a group of users (who worked on a paper using this dataset \cite{robot_paper}). They used the topics we found for a paper in which they also described the subjective topics analysis. In their paper, they looked at the topics and the words ranked in those topics and defined a topic name after checking the top words in each topic. This information can be seen in Figure \ref{fig:robo_topics}. With the help of our data they also analyzed which topics were important in which years, as shown in Figure \ref{fig:robo_heatmap}.\\
The results and the paper suggest that the algorithm found topics and combinations of words that make sense to scientists involved with robotics. We asked the authors and test persons (in total 5 persons) to score five simple questions with a scoring system of 1 (very bad) to 5 (very good):\\
\begin{enumerate}
    \item \emph{How do you grade the coherency of words in each single topic cluster (ignoring the time factor)?}
    \item \emph{How do you grade the coherency of the whole time series of the different topics? }
    \item \emph{How do you grade the ranking of words in the topics?}
    \item \emph{How do you grade the evolution of each of the topics over time? Do the changes in importance make sense to you?}
    \item \emph{Did the list of ranked words offer you an explanation for a topic's movement in a time series?}
\end{enumerate}
\begin{figure}
    \centering
    \includegraphics[width=.30\textwidth,keepaspectratio]{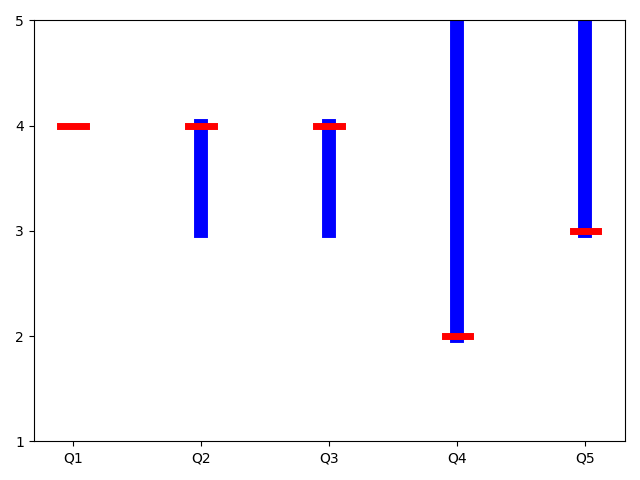}
    \caption{The distribution of answers for questions one to five. }
    \label{fig:question_overview}
\end{figure}
The range of answers for each question can be seen in Figure \ref{fig:question_overview}. The most answered number is shown by a vertical red line. The results indicate that users found that words put together in topics actually fit together and that the time series our greedy algorithm found made sense and showed the evolution of single topics. While the ranking of words in the topics mostly made sense to the test persons, they did not agree if the evolutions of topic made sense to them, with answer values ranging from "bad" to "very good." Finally, they stated that only sometimes the words in a topic explained why this topic was important or changed importance.\\

\subsubsection{Case Evaluation}\label{sec:case_evaluation}
\begin{figure}
    \centering
    \includegraphics[width=.60\textwidth,keepaspectratio]{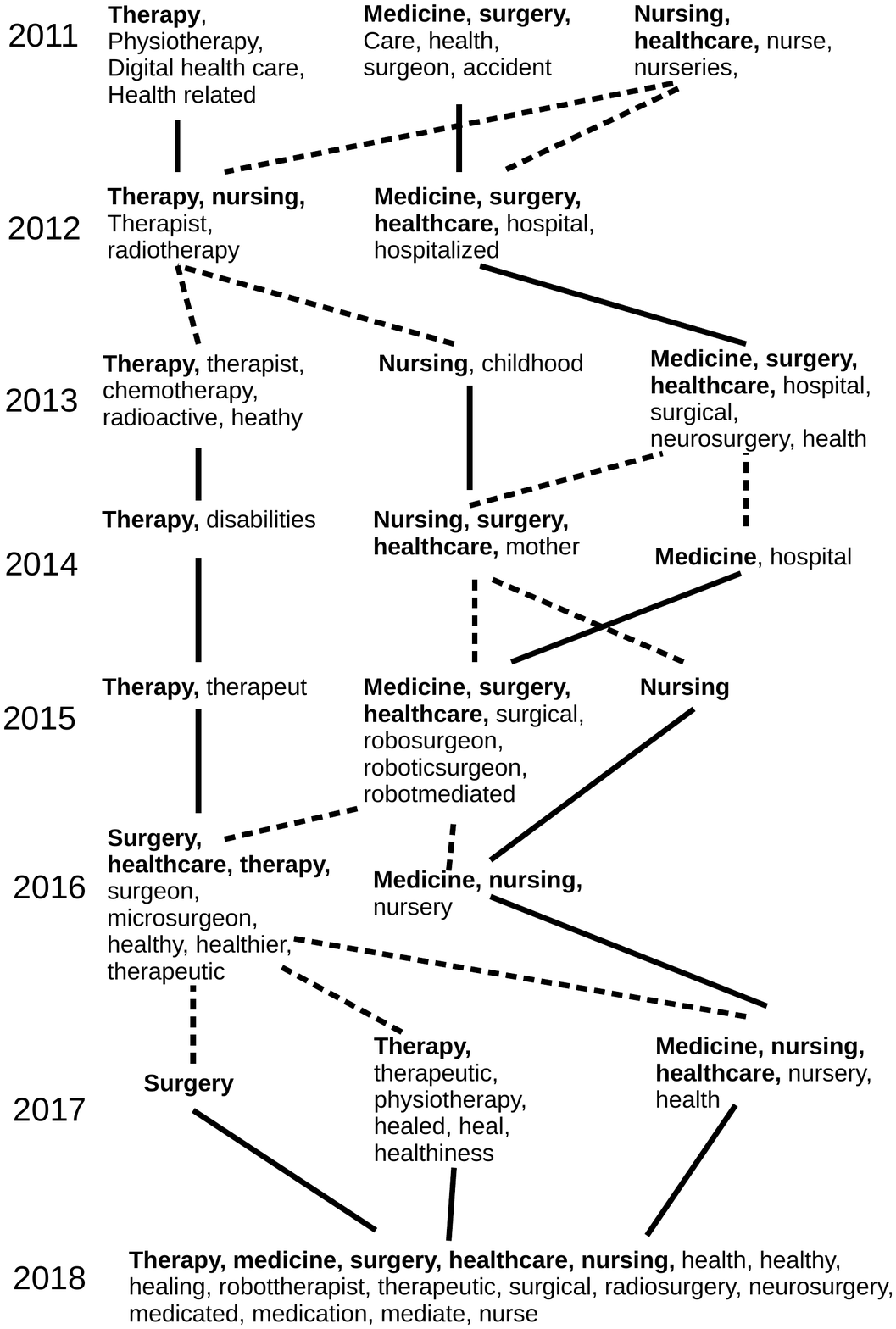}
    \caption{the evolution of healthcare regarding the domain robotics over time. Dashed lines show the splitting of topics. }
    \label{fig:timeline_robotics}
\end{figure}

\begin{figure}
    \centering
    \includegraphics[width=.60\textwidth,keepaspectratio]{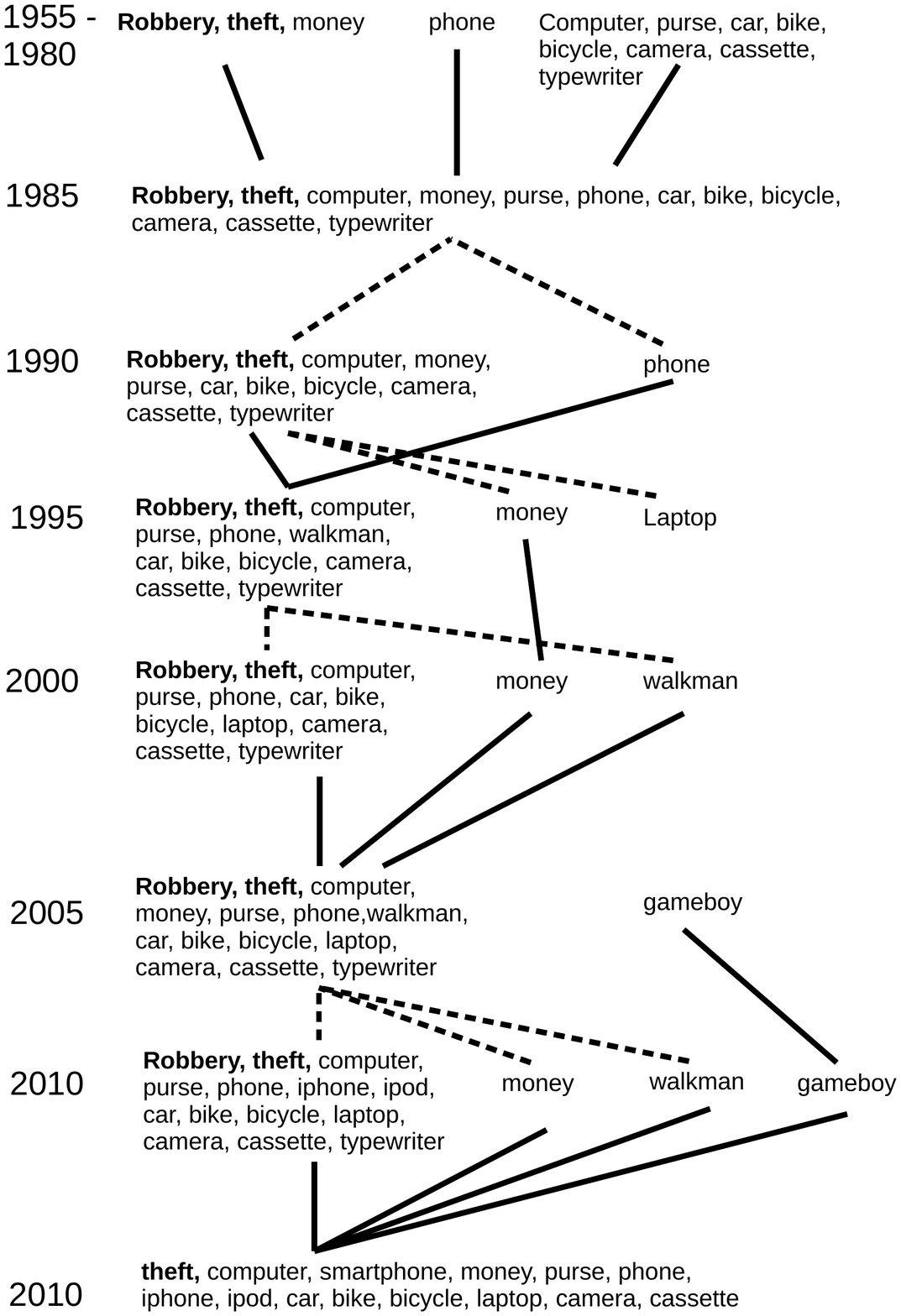}
    \caption{the evolution of theft in the Legal Dataset. Dashed lines show the splitting of topics. }
    \label{fig:timeline_legal}
\end{figure}

We chose two topics, one for each of the two datasets to show an example use case of evolving data.\\\\
\emph{Subjective Case Evaluation for the topic "Healthcare" from the Robotics Data Set:}\\
Figure \ref{fig:timeline_robotics} shows the evolution of the topic "healthcare" and what other words they are linked to. One thing visible is that due to the number of data increasing in 2018, the algorithm did not work as granular anymore and created one big healthcare-related topic instead of many smaller ones. From the words gathered from each topic, one can read what robotic development was used within what year, for example radiotherapy, physiotherapy or surgical assistance.\\\\
\emph{Subjective Case Evaluation for the topic "Theft" from the Legal Data Set:}\\
In the legal dataset, we looked at the specific topics containing words like "theft" or "robbery" to see what words these terms were mostly connected with in what year. This made it possible to see which objects were trending with the topic of theft in which years and track what was stolen more at what point in the timeline. The results can be seen in Figure \ref{fig:timeline_legal}.

\subsection{Results of the Quantitative Evaluation}
\subsubsection{Topic Coherency Evaluation:} An evaluation of topic models is usually based on estimated likelihood or perplexity. However, they are inadequate proxies for how topics are perceived by humans. 
According to Newman et al.~\cite{Newman_externalevaluation}, a topic has frequently some odd-words-out in the list of top words. This leads to the idea of a scoring model based on word association between pairs of words for the top-5 word pairs in a topic. For measuring word association, instead of using the collection itself, a large source of external text data should be used (in order to avoid reinforcing noise or unusual word statistics). PMI is seen as the measure of word association. Therefore, for considering human perception of topics we used pointwise mutual information (PMI) based on an external data source, Google 5-grams data.
\\
Using Google 5-gram data as an external source\footnote{We use the Google 5-gram data as the external data source as it supports a temporal comparison of words. The result obtained is similar to an evaluation using Wikipedia~\cite{Newman_externalevaluation}}, Momeni et al. \cite{Elaheh} counted
a co-occurrence of the topic's five highest ranked words in any of the 5-gram and scored a topic's coherence by averaging the PMI score of words. A higher average PMI score implies a more coherent topic. They assessed the average PMI score for 400 randomly selected topics (100 proposed by our approach for each dataset and 100 proposed by DTM for each dataset). They found that the median coherency of all topics proposed by their approach (median-Legal = $8.1$) is higher than the coherency of all topics proposed by the  DTM approach (median-Legal = $6.3$). \\
Furthermore, they selected 25 topics and asked five human subjects (who can speak both languages) to score each of the 50 topics on a three-point scale where 3=``useful'' and 1=``useless''. The human scoring of these topics has a high inter-rater reliability measurement using Fleiss' kappa score ($0.71$).  
Finally, we see broad agreement between the average PMI score
and the human scoring. The correlation between
the average PMI score and the mean human score is $\rho=0.74$ for the Legal data (we define correlation $\rho$ as the Pearson correlation coefficient).\\
To investigate how our optimization approches improve the coherence of words in topics, we used the same PMI metric as \cite{Elaheh} did. We calculated our PMI scores in the following way:
\begin{enumerate}
    \item Choose 10 random clusters
    \item Rank the top 10 words for each cluster based on the words TF-IDF value
    \item Create word pairs for each cluster's top 10 words where each word is paired with each of the top 10 words
    \item Search the whole google 5-gram dataset for 5 grams in which a word pair occurs together (order and capitalization do not matter). Additionally, count how often every single word appears in the google ngram dataset
    \item Calculate the PMI score for each topic using the following formulas \cite{Newman_externalevaluation}:
    $$
    PMI-Score(w) = median\{PMI(w_i , w_j ), ij \epsilon 1 . . . 10\}
    $$
    and
    $$
    PMI(w_i, w_j) = log\frac{p(w_i, w_j)}{p(w_i)p(w_j)}
    $$
    where $w_i$ and $w_j$ describes one word from a topic
\end{enumerate}
 The average PMI for the legal data set with sortable words was 9.4, while the unsorted data PMI value was 8.1 \cite{Elaheh}. This shows that we could improve cohrency, using our optimization approach and significantly reduce unwanted elements with our reduction techniques. It should be noted that the unsorted data also resulted in fewer values being found in the Google ngram data. This may be because the unsorted randomness resulted in looking for unused word combinations.

\subsubsection{Execution Time}
\label{sec:execution:time}
\begin{figure*}
 \centering
\includegraphics[width=.95\textwidth]{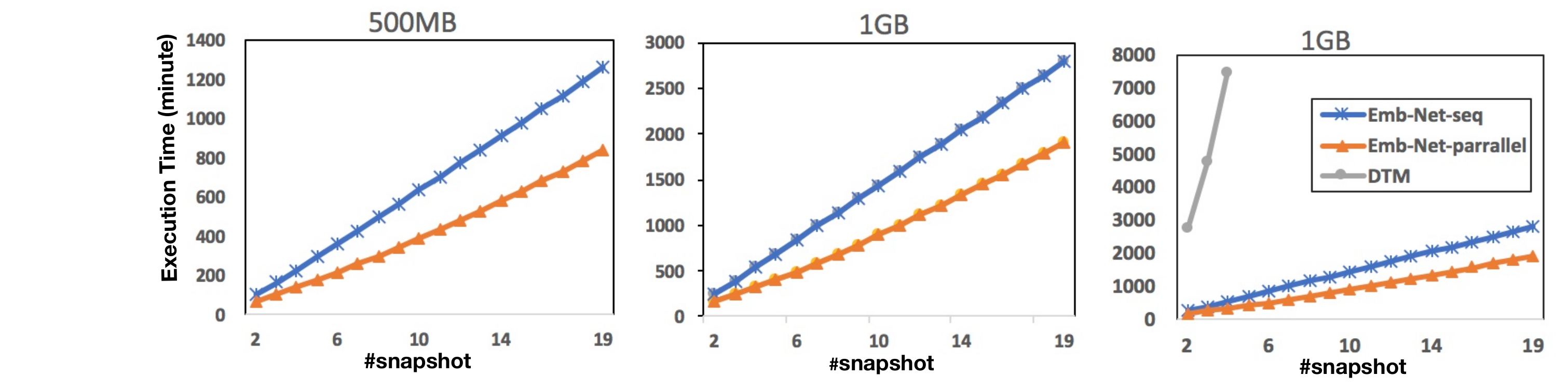}
\caption{\small{Execution time for different sizes and numbers of snapshots in 2 various set ups: sequential and parallel\cite{Elaheh}}}
\label{fig:excDTM}
\end{figure*}

Finally, we investigate how efficient our approach performs in different set ups. As this approach can perform each step independently on each snapshot, it can run the whole process in parallel (multi-thread processing) to reduce the running time. 
\\
More precisely, Figure \ref{fig:excDTM} from \cite{Elaheh} shows execution time for different sizes of snapshots. Size of snapshots in both datasets range between 100MB-5GB (smaller size in earlier snapshots). This figure shows execution time for average snapshot size of 500MB and 1GB. The Figure also shows the effect of different numbers of snapshots in different set-ups: sequential (run the process sequentially for each snapshot) and parallel (run the process in parallel for all snapshots) running. Finally, the last plot shows comparison of the execution time of DTM approach with the execution time of this approach. As execution time of DTM was long (almost 5 days for four snapshots with average size of 1GB) they just trained the models for the latest 3 snapshots of each dataset. They executed the process on a machine (i7 core processor and 64GB RAM space). Figure \ref{fig:excDTM} clearly demonstrates the scalability of their approach for detecting and modeling evolution of dynamic topics in a large-scale text corpora. It executes for an average snapshot size of 1GB for all 19 snapshots around 30 hours in the parallel set up. 

\begin{figure*}[h]
\centering
\includegraphics[width=0.70\textwidth]{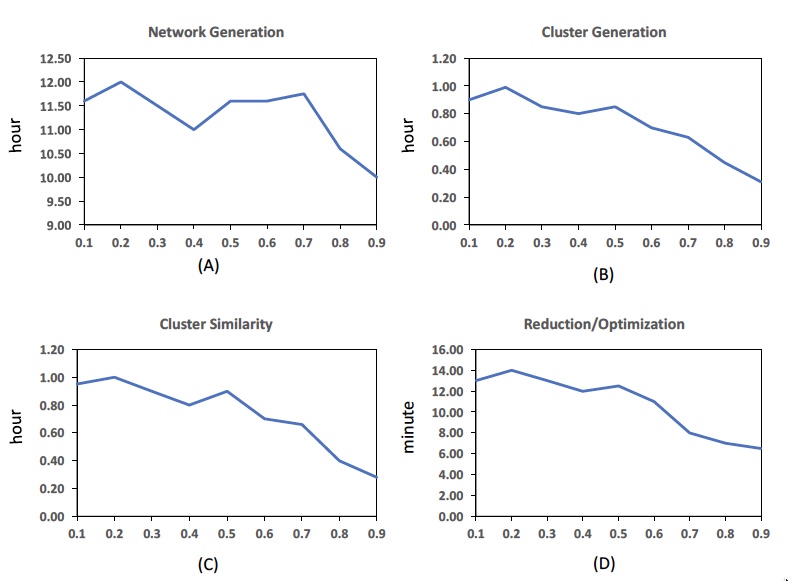}
\caption{Plots of execution time for different parts of the system considering similarity threshold. Higher threshold results in a semantic network with lower density.}
\label{fig:evaluation_time}
\end{figure*}

Furthermore, in order to obtain in-depth understanding about the effect of different parts of the system and our proposed optimization approaches on execution time, we evaluated the execution time of each part separately considering various setups of thresholds.
\\\\
\textbf{Data aggregation and model generation.} The only part of the program that we only need to run once is the model generation. It ran for 2 hours, 21 minutes, and 13 seconds. This period includes reading the data from a source (such as CSV files), tokenizing them, and creating the word2vec models.\\This value is mainly dependent on the number of texts needed to be read and tokenized, and the word2vec model parameters chosen.\\
\\
\textbf{Network Generation.}
The times for the network generation can be seen in part A of Figure \ref{fig:evaluation_time}. As the figure shows, the network generation is a task that takes considerable time. In fact, it is the part of the whole algorithm that takes the longest. This is most likely due to the nature of the network generation, where the algorithm loops over every word and creates connections to the top N most similar words. \\
The execution time of the network generation, like the model generation, is mainly dependent on the size of the input data. The more words the model vocabulary has, the more words need to be analyzed as we are analyzing the top 10 similar words for each word in the vocabulary.\\
\textbf{Cluster Generation.}
Part B of Figure \ref{fig:evaluation_time} shows the distribution of execution time for each word similarity threshold. At the more restrictive thresholds (0.8, 0.9), there is a clear trend visible that the cluster generation took less time than with a lower threshold. The reason here is, as we can also see in the output files, that such a high word similarity threshold results in very small networks. This also means that the Louvain clustering algorithm has less data to consider and fewer clusters are created.\\The execution time for the cluster generation is mainly dependent on the size of the generated networks as the cluster generation has to analyze the nodes and edges in each network.\\
\textbf{Cluster Similarity Calculation}
When analyzing the similarity calculation, which is shown in part C of Figure \ref{fig:evaluation_time}, one can see a similar pattern to what was already the case with the cluster generation. With the last thresholds, a big decrease in computation time can be seen as there are fewer clusters to consider for comparison.\\The similarity calculation is dependent on the number of clusters generated and their size.\\
\textbf{Word \& cluster Reduction}
Part D of Figure \ref{fig:evaluation_time} shows the evolution of running times for the cluster and word reduction that are done in the same step. For readability and space reasons, we used the mean time for each word similarity threshold value. This also results in the chart being comparable to the previous evaluation charts about execution time.
The similarity to previous charts about execution time is apparent. The same logic can also be applied in this case, where the mean time of the 0.9 word similarity threshold is significantly lower again. The rest of the chart follows the same patterns as the previous chart. It should be noted that the reduction step, in general, took less time than all the above steps.\\
The execution time of this part of the algorithm is again only dependent on the number of clusters and their size as we are analyzing each cluster independently. The estimation of the creation of time series is not easy as the execution time is dependent on the length of the series, and this cannot be defined beforehand.

\section{Conclusion}
Taking into consideration the previous research on modeling the evolution of trending topics by \cite{Elaheh}, two challenges remain to enable better automatic topics modeling and further processing. On the one hand, their algorithm resulted in too many clusters that needed to be filtered because they were not essential or helpful. On the other hand, the words in each topic needed filtering as well as ranking in order to make sense of them. 
\\
We took the algorithm presented in \cite{Elaheh} and, besides significant speed improvements, implemented a network analyzing component to obtain characteristics of the topics. We implemented calculations and filtering based on known network analysis metrics such as a topic's centrality, density, and cluster frequency value. We then tested different combinations of these values to find an appropriate solution that works for big datasets and can filter topics.
\\
Additionally, we added the k-core decomposition method to obtain the main words that are in the conversation of a topic. Also, to show the more original words in a topic, the calculation of TFIDF-values has been added that ranks words inside a cluster and identifies the most meaningful and novel words. On this basis, we can add a filter to reduce the number of words inside a topic. 
\\
The final algorithm provides a method, first, to filter clusters by their centrality and cluster frequency values and, second, to filter and rank words to keep only essential words. We evaluated the results in different set-ups, including a subjective user evaluation and an objective topic coherency evaluation of the filtered topics. We were able to show the effectiveness and efficiency of our enhancement modules for the whole system.
\\
A comparison of our qualitative and quantitative evaluation shows that these two mainly agree. Our test persons mostly found the proposed topics interesting and agreed that the coherency of the resulting topics and the coherency of a full timeline of a topic made sense to them and was usable for further processing or work. They also agreed that the ranking of words in a topic was helpful and that this ranking could make a future automatic analysis possible. Additionally, it was possible to create case studies for two randomly chosen topics, healthcare (Figure \ref{fig:timeline_robotics}) and theft (Figure\ref{fig:timeline_legal}), and analyze, how a topic evolved over time concerning which words were important at which timestamp and therefore able to analyze a certain topic.\\
The qualitative evaluation agreed with the test persons. It showed that we had a definite improvement concerning the pairwise mutual information score compared to the score in the state-of-the-art work in \cite{Elaheh} and agreed with our test persons that the resulting topics were coherent. Additionally, our run time changes to the algorithm result in a dramatic improvement of the execution time compared to \cite{Elaheh}, even though new features were added.\\
Considering the results of the evaluation, a good basis has been prepared to make future processing and enhancement, especially for the end-user, possible.
\subsection{Future Work}
\emph{Enhancing the Time Series Analysis}
Currently, this algorithm uses a greedy method to find a cluster's timeline and merges topics. This means that it assumes that the next cluster in a timeline is always the one with the biggest similarity, where the similarity is defined by the number of words that occur in both clusters, normalized by the number of nodes. If multiple topics have the most similar topic in the next timestamp, a merge is created. In the future, different ways for finding merges and time series could be tested to find a better way than the greedy algorithm.\\
\emph{Adding Prediction to predict Events based on past Events}
A future extension could be able to predict events and topic evolution over time. This would enable us to predict when a topic "dies" based on past events and when a topic could become more important again based on centrality, density, and frequency values.\\
\emph{Adding a parameter-tuning Algorithm}
The reality of this algorithm is that many parameters need to be set. Additionally, the parameters are very high level. Switching to easier, human understandable values would make it easier to use this algorithm. An example would be that the percentage thresholds for the reduction could be replaced by more user-friendly and easier to understand parameters, such as "strong" which shows how many topics and words the algorithm should filter. The algorithm could then finetune itself based on these settings, making it more user-friendly.
The addition of a sensitivity slider to a front-end application could be a help to omit the complexity of thresholds. In this case, the algorithm would then try to tune the filtering based on the sensitivity.\\
\emph{Visualizing merges and changes of importance}
The key to such an algorithm is a good, thought-through user interface that lets a user explore the data. A user interface should offer users a choice of necessary settings: setting thresholds themselves and setting the year they look at. There should also be an animated component offering an animation for the topics describing how it rises and falls in importance over time. Adding an overview of the top-ranked words would also be important. Additionally, a user interface that lets a user search for specific topics and its top words could be helpful to design use cases like ours use cases.\\
\emph{Normalizing values for enhanced filtering}
Improving filtering could be done by normalizing the centrality values. This would be needed to create better and automatic support for different domains. For our legal domains, centrality values always seemed to be very small.  Hence, we had to adapt our thresholds by hand. By normalizing them, this algorithm could be enhanced to work better automatically with different domains.

\small
\bibliographystyle{aaai}
\bibliography{ref}

\begin{thebibliography}{}

\bibitem[\protect\citeauthoryear{Aiello \bgroup et al\mbox.\egroup
  }{2013}]{pres3}
Aiello, L.~M.; Petkos, G.; Martin, C.; Corney, D.; Papadopoulos, S.; Skraba,
  R.; Göker, A.; Kompatsiaris, I.; and Jaimes, A.
\newblock 2013.
\newblock Sensing trending topics in twitter.
\newblock {\em IEEE Transactions on Multimedia} 15(6):1268--1282.

\bibitem[\protect\citeauthoryear{Anagnostopoulos \bgroup et al\mbox.\egroup
  }{2016}]{Aris:2016}
Anagnostopoulos, A.; Lacki, J.; Lattanzi, S.; Leonardi, S.; and Mahdian, M.
\newblock 2016.
\newblock Community detection on evolving graphs.
\newblock In {\em Advances in Neural Information Processing Systems 29: Annual
  Conference on Neural Information Processing Systems 2016, December 5-10,
  2016, Barcelona, Spain},  3522--3530.

\bibitem[\protect\citeauthoryear{Blei and Lafferty}{2006}]{Blei:2006}
Blei, D.~M., and Lafferty, J.~D.
\newblock 2006.
\newblock Dynamic topic models.
\newblock In {\em Proceedings of the 23rd International Conference on Machine
  Learning}, ICML '06,  113--120.
\newblock New York, NY, USA: ACM.

\bibitem[\protect\citeauthoryear{\.{I}lhan and
  \"{O}\u{g}\"{u}d\"{u}c\"{u}}{2015}]{Ilhan:2015}
\.{I}lhan, N., and \"{O}\u{g}\"{u}d\"{u}c\"{u}, c.~G.
\newblock 2015.
\newblock Predicting community evolution based on time series modeling.
\newblock In {\em Proceedings of the 2015 IEEE/ACM International Conference on
  Advances in Social Networks Analysis and Mining 2015}, ASONAM '15,
  1509--1516.
\newblock New York, NY, USA: ACM.

\bibitem[\protect\citeauthoryear{Javaheri \bgroup et al\mbox.\egroup
  }{2019}]{robot_paper}
Javaheri, A.; Moghadamnejad, N.; Keshavarz, H.; Javaheri, E.; Dobbins, C.;
  Momeni, E.; and Rawassizadeh, R.
\newblock 2019.
\newblock Public vs media opinion on robots.
\newblock {\em CoRR} abs/1905.01615.

\bibitem[\protect\citeauthoryear{Mikolov \bgroup et al\mbox.\egroup
  }{2013a}]{Mikolov:2013}
Mikolov, T.; Chen, K.; Corrado, G.; and Dean, J.
\newblock 2013a.
\newblock Efficient estimation of word representations in vector space.
\newblock {\em CoRR} abs/1301.3781.

\bibitem[\protect\citeauthoryear{Mikolov \bgroup et al\mbox.\egroup
  }{2013b}]{Mikolov2:2013}
Mikolov, T.; Sutskever, I.; Chen, K.; Corrado, G.; and Dean, J.
\newblock 2013b.
\newblock Distributed representations of words and phrases and their
  compositionality.
\newblock In {\em Proceedings of the 26th International Conference on Neural
  Information Processing Systems - Volume 2}, NIPS'13,  3111--3119.
\newblock USA: Curran Associates Inc.

\bibitem[\protect\citeauthoryear{Momeni \bgroup et al\mbox.\egroup
  }{2018}]{Elaheh}
Momeni, E.; Karunasekera, S.; Goyal, P.; and Lerman, K.
\newblock 2018.
\newblock Modelling evolution of topics in large-scale text corpora.
\newblock In {\em International AAAI Conference on Weblogs and Social Media
  (ICWSM 2018)}.

\bibitem[\protect\citeauthoryear{Newman, Karimi, and
  Cavedon}{2009}]{Newman_externalevaluation}
Newman, D.; Karimi, S.; and Cavedon, L.
\newblock 2009.
\newblock External evaluation of topic models.
\newblock In {\em in Australasian Doc. Comp. Symp., 2009},  11--18.

\bibitem[\protect\citeauthoryear{Saquib and Ali}{2017}]{Pres1}
Saquib, S., and Ali, R.
\newblock 2017.
\newblock Understanding dynamics of trending topics in twitter.
\newblock In {\em 2017 International Conference on Computing, Communication and
  Automation (ICCCA)},  98--103.

\bibitem[\protect\citeauthoryear{Stokman}{2001}]{STOKMAN200110509}
Stokman, F.
\newblock 2001.
\newblock Networks: Social.
\newblock In Smelser, N.~J., and Baltes, P.~B., eds., {\em International
  Encyclopedia of the Social \& Behavioral Sciences}. Oxford: Pergamon.
\newblock  10509 -- 10514.

\bibitem[\protect\citeauthoryear{Tantipathananandh, Berger-Wolf, and
  Kempe}{2007}]{Tantipathananandh:2007}
Tantipathananandh, C.; Berger-Wolf, T.; and Kempe, D.
\newblock 2007.
\newblock A framework for community identification in dynamic social networks.
\newblock In {\em Proceedings of the 13th ACM SIGKDD International Conference
  on Knowledge Discovery and Data Mining}, KDD '07,  717--726.
\newblock New York, NY, USA: ACM.

\bibitem[\protect\citeauthoryear{Zubiaga \bgroup et al\mbox.\egroup
  }{2011}]{Pres2}
Zubiaga, A.; Spina, D.; Fresno, V.; and Mart\'{\i}nez, R.
\newblock 2011.
\newblock Classifying trending topics: A typology of conversation triggers on
  twitter.
\newblock In {\em Proceedings of the 20th ACM International Conference on
  Information and Knowledge Management}, CIKM '11,  2461--2464.
\newblock New York, NY, USA: ACM.

\end{thebibliography}

\end{document}